\pdfoutput=1

\documentclass[11pt]{article}
\usepackage{acl}
\usepackage{cjhebrew}

\usepackage{times}
\usepackage{latexsym}

\usepackage[T1]{fontenc}

\usepackage[utf8]{inputenc}

\usepackage{microtype}

\usepackage{import}
\usepackage{fancyvrb}
\usepackage{xspace} 

\usepackage{enumitem}
\usepackage{subcaption}

\newcommand{\ptb}{PTB\xspace}

\newcommand{\pos}{POS\xspace}

\newcommand{\glove}{GloVe\xspace}

\newcommand{\ytb}{PPCHY\xspace}
\newcommand{\ybc}{YBC\xspace}

\newcommand{\ocr}{OCR\xspace}
\newcommand{\html}{html\xspace}
\newcommand{\ascii}{ASCII\xspace}
\newcommand{\svo}{SYO\xspace}
\newcommand{\syo}{SYO\xspace}
\newcommand{\nlp}{NLP\xspace}
\newcommand{\fasttext}{fastText\xspace}
\newcommand{\xlmrobertabase}{XLM-roberta-base\xspace}
\newcommand{\flair}{Flair\xspace}

\title{A Part-of-Speech Tagger for Yiddish}

\author{Seth Kulick \and Neville Ryant \\
Linguistic Data Consortium \\
University of Pennsylvania \\
\texttt{\{skulick,nryant\}@ldc.upenn.edu} \\
\And 
Beatrice Santorini \\
Department of Linguistics \\
University of Pennsylvania \\
\texttt{beatrice@sas.upenn.edu} \\
\AND
Joel Wallenberg \\
Department of Language\\
and Linguistic Science \\
University of York \\
\texttt{joel.wallenberg@york.ac.uk}
\And
Assaf Urieli \\
Joliciel Informatique \\
Foix, France \\
\texttt{assaf@joli-ciel.com}
}

\date{}

\begin{document}
\maketitle

\begin{abstract}
We describe the construction and evaluation of a part-of-speech 
tagger for Yiddish. This is the first step in a larger project of automatically assigning part-of-speech tags and syntactic structure to Yiddish text for purposes of linguistic research.  We combine two resources for the current work - an 80K-word subset of the Penn Parsed Corpus of Historical Yiddish (\ytb) \citep{yiddishtb} 
and 650 million words of OCR'd Yiddish text from the Yiddish Book Center (\ybc). 
Yiddish orthography in the \ybc corpus has many spelling inconsistencies, and we present some evidence that even simple non-contextualized embeddings trained on \ybc  are able to capture the relationships among spelling variants without the need to first ``standardize'' the corpus. 
 We also use \ybc for continued pretraining of contexualized embeddings, which are then integrated into a  tagger model trained and evaluated on the \ytb. 
  We evaluate the tagger performance on a 10-fold cross-validation split, showing that the use of the \ybc text for the contextualized embeddings improves tagger
performance.   We conclude by discussing some next steps, including the need for additional annotated training and test data.
\end{abstract}

\section{Introduction}
Treebanks - text corpora annotated for part-of-speech (POS) and syntactic information - have been important resources for both natural language processing (\nlp) and research on language change. 
For NLP, corpora such as the Penn Treebank (\ptb) \citep{marcus-etal-1993-building}, consisting of about 1 million words of modern English 
text, have been crucial for training machine learning models intended to automatically annotate new text with POS and syntactic information. 

For research on language change, a family of treebanks of
historical Eng\-lish \citep{ycoe, ppcme2, pceec, ppcmbe}  and other languages \citep{icepahc, tycho-brahe, mcvf, ppchf}, with a shared annotation philosophy and similar guidelines across languages, has formed the basis for reproducible studies of language change \citep{kroch-taylor-ringe, ecay, wallenberg-2016, galves-20, wallenberg-et-al-2021}.   

The Penn Parsed Corpus of Historical Yiddish (\ytb) \citep{yiddishtb}, developed mostly in the late 1990s, is an early example of such a treebank developed for linguistic research.  It consists of about 200,000 words of Yiddish dating from the 15th to 20th centuries, annotated with \pos tags and syntactic trees. The corpus grew out of a collection of linguistically relevant examples for research on the changing syntax of subordinate clauses in the history of Yiddish \cite{santorini1989diss, santorini1992variation, santorini1993rate}.

While the PPCHY contains a relatively small  amount of treebanked Yiddish text, there is a large amount of unannotated Yiddish text available.  In particular, the Yiddish Book Center (\ybc) has applied Yiddish-specific Optical Character Recognition
(OCR)\footnote{\url{ https://github.com/urieli/jochre}}
to their book collection, resulting in about 650 million words of Yiddish text \citep{Markey,urieli2013,kutzik}.

We describe here the development of a \pos-tagger bringing these resources together -  using the \ytb as training and evaluation material, and using the \ybc corpus to train word embeddings, which encode distributional similarities among words and which we integrate into the tagger training.  The use of embeddings can improve a model's performance beyond the immediate annotated training data and has resulted in great advances in \nlp  over the last few years.  We also present evidence that the embeddings implicitly encode information on spelling variations, thereby allowing us to bypass a step of standardizing the large number of spelling variants in the \ybc texts.

This \pos tagger is the first step in a larger project of (1) expanding \ytb to one million words of manually annotated text, and (2) training a \pos tagger and syntactic parser for Yiddish to automatically annotate a large and historically diverse corpus of Yiddish, including the \ybc itself.  This would allow for a next phase in corpus-based diachronic research on Yiddish by increasing the amount of annotated text beyond what could be manually annotated. We also hope that the steps in this work can result in additional search capabilities on the \ybc corpus (e.g., by \pos tags), and possibly the identification of orthographic and morphological variation within the texts, including candidates for OCR post-processing correction. 

The goal of the larger project is analogous to, and inspired by, current work with Early Modern English (1500-1700)
\citep{kulick-etal-2022-penn,kulick-etal-2022-parsing,kulick-etal-2023-parsing}
 that aims to massively increase the amount of annotated text for research on the history of English. Our approach to orthographic variation in the Yiddish text is also inspired by this work, which similarly faces a great deal of inconsistent spellings in English text of that time period.

In Sections \ref{sec:orthography} and \ref{sec:previous} we present some background to the present work.  Sections \ref{sec:ybc} and \ref{sec:ytb} discuss the \ybc corpus and the \ytb.  Sections \ref{sec:common} and \ref{sec:conversion} discuss the problem of needing a common representation for the two sources and our solution to that problem.  Section \ref{sec:embeddings} describes the creation of the embeddings using the \ybc.
Sections \ref{sec:model} and \ref{sec:results} discuss the \pos tagger model\footnote{The software for the resulting \pos tagger is at \url{https://github.com/skulick/yiddishtag}.} and the results, and Section \ref{sec:conclusion} is the conclusion.

\section{Yiddish Orthography}
\label{sec:orthography}

This section summarizes some aspects of Yiddish orthography that are referred to in following sections. For more details on these issues, see \citet[pp. 667-670]{kahn}, \citet[pp. 46-52, 301-303]{jacobs}, and \citet{gold}.

Yiddish is in general written with Hebrew characters, although the representation of vowels and the use of diacritics is significantly different than in Hebrew. Yiddish can be roughly considered to consist of German and Hebrew/Aramaic components, with two different spelling systems, and each present important aspects to be considered for \nlp work.

\paragraph{Spelling inconsistency for the German component}
This component is considered ``phonetic'' in that there is a mapping from the orthography to the pronunciation. However, the spelling has varied in quite a few ways over time, sometimes reflecting efforts to follow German spelling (``daytshmerish orthography'') or reflecting dialect pronunciations.  In Section \ref{sec:embeddings} we give some examples of these spelling variations, and show how the word embeddings reflect them.

There have been proposals for a standardized orthographic system, in particular by the YIVO Institute for Jewish Research in 1936, 
which we will refer to as ``Standard YIVO  Orthography'' (\syo).  While this has in some respects come to be the standard for contemporary Yiddish, most of the text in the \ybc corpus predates this proposal, and even subsequent text doesn't necessarily follow it.  

\paragraph{``Non-phonetic'' component}  
While the spelling of words from the Hebrew/Aramaic component is less variable, they are represented by what is in essence a second orthographic system. For example, some letters only appear in these words.  The frequency of words from this component  can vary considerably depending on the material.\footnote{\citet[p. 66]{weinreich} writes that it can sometimes exceed 15 percent.}  For the tagger described here, their identification is important because the tagset has tags for Hebrew words used as particles and for quoted Hebrew text.

\paragraph{Romanization} There is a relatively standard transliteration from the Yiddish script, particularly from \syo orthography, to Latin letters.  For the phonetic component, the mapping from the Yiddish script to the romanization and back is fairly straightforward (modulo spelling differences in the Yiddish script).  This not always the case, though.  For example, \textit{alts} can be \<'al.s> `all' or \<'al.ts> (a form of `old').  However, for the non-phonetic component, the mapping is not simple and in effect requires listing each such case.\footnote{\citet{saleva-2020-multi} notes that the Hebraic words were problematic for the romanized-to-\syo transliteration model, since the model incorrectly applied the spelling rules it had learned to those words as well.}  There are also ambiguities where different words from the two components may be transliterated as the same romanization. For example, the romanization \textit{shem} can be the ``non-phonetic'' noun \</sM> `reputation' or the ``phonetic'' verbal stem  \</s`M> `be shy'.

Such ambiguities are an important concern for this work, since 
the \ytb files discussed in Section \ref{sec:ytb}, unlike the \ybc texts, are romanized and must be converted to Yiddish script in order to prepare the training and evaluation material for the tagger, as discussed in Sections \ref{sec:common} and \ref{sec:conversion}.

\section{Related Work}
\label{sec:previous}

\citet{kirjanov}, \citet{blum2015}, and \citet{saleva-2020-multi} all discuss the problem of normalizing Yiddish text to a standard form.  \citet{blum2015} experiments with a number of approaches to convert Yiddish text to \syo (including \citet{kirjanov}).   It is noteworthy that the training and evaluation data for this work are an earlier form of the OCR output used for \ybc, although much smaller (16 documents and 37,902 tokens), with OCR mistakes manually excluded.  The work used a list of standardized forms for all the words in the texts, experimenting with approaches that match a variant form to the corresponding standardized form in the list. 

\citet{saleva-2020-multi} uses a corpus of Yiddish nouns scraped off Wiktionary and consisting of 2,750 word forms to create transliteration models from \syo to the romanized form, from the romanized form to \syo, and from the ``Chasidic'' form of the Yiddish script to \syo, where the former is missing the diacritics in the latter.

We view these works as complementary to the current work. The work described below involves 650 million words of text which is internally inconsistent between different orthographic representations, along with the inevitable \ocr errors, and we do not have a list of the standardized forms of all the words in the \ybc corpus.   However, it is possible that continued work on the \ybc corpus will further development of transliteration models. 

There are two sources of word embeddings for Yiddish currently available.  Non-contextualized embeddings for Yiddish have been created by \citet{grave-etal-2018-learning} as part of the \fasttext word embeddings for 157 languages.  The \fasttext embeddings were in general created using Wikipedia and CommonCrawl, although the specific sources or quantity were  the Yiddish embeddings are not clear.  The multilingual \xlmrobertabase contextualized embeddings \citep{conneau-etal-2020-unsupervised} include training on 34M Yiddish tokens (0.3 GB).   We return to the use of these embeddings in Section \ref{sec:embeddings:con}.

\section{The Yiddish Book Center Corpus}
\label{sec:ybc}
We describe here the main steps we took to generate the \ybc corpus. See Appendix \ref{app:ybc} for more details. 

\paragraph{Downloading and text extraction}  To assemble the \ybc corpus, we downloaded 9,925 \ocr \html  files from the Yiddish Book Center site, performed some simple character normalization, extracted the \ocr'd Yiddish text from the files,  and filtered out 120 files due to rare characters, leaving 9,805 files to work with.

\paragraph{Conversion to \ascii}  The files were in the Unicode representation of the Yiddish alphabet.  For ease of processing, we preferred to work with a left-to-right version of the script within strict ASCII. We therefore
defined a bidirectional mapping between the Unicode and \ascii and converted the \ybc text to this \ascii representation. We omit the details here of the definition of the notational variant, except to say that we tested it by converting all of the download text from the Unicode 
to \ascii and back again.

\paragraph{Tokenization and sentence segmentation} 
With the text now in a more convenient representation, we also did some simple tokenization, separating out most cases of punctuation (e.g., ``\<bwK\textrm{.}>'' becomes ``\<bwK \textrm{.}>'').\footnote{We did not attempt at this stage to account for abbreviations with a period, and currently the period is incorrectly separated in abbreviations, which also affects the sentence segmentation.}  
We did not, however, split words on apostrophes,  given its frequent and inconsistent use in Yiddish.  
For example, the word  \<'mt> {\it emes} `truth', with an inflectional
ending {\it n}, appears 27,129 times as \<'mtn> and 12,743 times as \<'mt\textrm{'}n>. 

We split the lines into sentences based on the presence of period, question mark, or exclamation mark. 

\paragraph{Result}
This process resulted in 9,805 
files with 653,326,190 whitespace-delimited tokens, in our ASCII equivalent of the Unicode Yiddish script.\footnote{Some of these tokens are punctuation marks, due to the tokenization process.  In what follows we will use the terms token and word interchangeably.} There were 10,642,884 distinct tokens.   In Appendix \ref{app:aspects} we discuss some aspects of how our use of the \ybc corpus might be improved, in particular with respect to the \ocr errors that will inevitably occur in uncorrected \ocr of a corpus of this size.

\section{The Penn Parsed Corpus of Historical Yiddish}
\label{sec:ytb}

As mentioned in the introduction, the \ytb contains text from the 15th to 20th centuries.   While most of the files contain varying amounts of running text, in some cases containing only subordinate clauses (because of the original research question motivating the construction of the treebank), the largest contribution comes from two 20th-century texts, \citet{grinefelder} (15,906 tokens) and \citet{olsvanger} (67,551 tokens). These are the two files from the \ytb that we use in the work reported here.\footnote{These are the files 
1910e-grine-felder and 1947e-royte-pomerantsn in the treebank.  The latter file was also used for a study of discourse functions of syntactic form in Yiddish \citep{prince1993discourse}.}  

An example tree from the \ytb is shown in (1) for the sentence (2), from \citet{grinefelder}.

\begin{minipage}{\linewidth}
\noindent (1)
{\small 
\begin{Verbatim}[samepage=true]
(IP-MAT
  (META (NPR rokhl:))
  (NP-SBJ (NPR elkone))
  (PUNC ,)
  (CP-QUE-MAT-PRN 
        (IP-SUB (VBF meyns@)
                (NP-SBJ (PRO @tu))))
  (PUNC ,)
  (VLF volt)
  (NEG nisht)
  (VB veln)
  (NP-ACC (NPR hersh-bern))
  (PP (P far)
      (NP (D an) (N eydem)))
  (PUNC ?))
\end{Verbatim}
}
\end{minipage}

\noindent (2) 

\begin{cjhebrew}
r.hl; 'lqnh \textrm{,} myyns.tw \textrm{,} ww'Al.t ny/s.t ww`ln h`r/s--b`rn
p'ar 'an 'yyd`m \textrm{?}
\end{cjhebrew}

(Rokhl: Elkone, do you think, would not want Hersh-Ber for a son-in-law?)
{\ }

\paragraph{Text representation}
One of the reasons we focus on the two files mentioned is that they use a romanization that mostly corresponds to the \syo (which is not always true for the older Yiddish text).  This is evident in the representation of the words in (1), which also contains an example of the slight modification from the usual romanized form, in that 
what are usually written as single words are sometimes split apart for purposes of the \pos and syntactic annotation.
Specifically, \<myyns.tw>  {\it meynstu} 
is written as two separate tokens {\it meyns} and {\it tu}, with the split indicated by the trailing and leading "at" signs. This allows {\it tu} to be annotated as the pronoun subject, while {\it meyns} is the finite verb.\footnote{The verb is actually {\it meynst}, but the {\it t} is in effect shared between the two separated tokens.}  Other common cases of this tokenization in the \ytb concern the separation of stressed
verbal prefixes
and contractions with an apostrophe, such as {\it s'iz}, which are split after  the apostrophe.

\paragraph{Syntactic and \pos annotation}
The text in the \ytb has been annotated in a similar style to that of other treebanks for historical research, such as the Penn Parsed Corpus of Historical English \cite{ppche} and the Icelandic Parsed Historical Corpus \citep{icepahc}. 

Without going into full detail, the tree shows the usage of the \pos tags NPR (proper noun), VBF (ordinary finite verb), PRO (pronoun), VLF (finite form of VOLN), VB (infinitive), P (preposition), N (noun), NEG (negation), and PUNC (punctuation).  The syntactic annotation on top of the \pos tags also shows that {\it meyns tu} is a parenthetical question, that {\it elkone} is the subject, {\it hersh-bern} is the accusative object, and {\it far an eydem} is a prepositional phrase.\footnote{For more details, the general annotation style is at
\url{https://www.ling.upenn.edu/~beatrice/annotation/index.html}.  See \citet{yiddishtb} for Yiddish-specific details.}

\section{The Need for a Common Representation}
\label{sec:common}
The two main resources for this work are, to this point, in different representations. 
The \ybc, discussed in Section \ref{sec:ybc}, is in Yiddish script, while  
the \ytb, discussed in Section \ref{sec:ytb}, is in a romanized form, with some whitespace-delimited tokens split into two.
We need to have a common representation for the two, in order to allow the embeddings trained on the \ybc to be used in the model trained on the \ytb, so that it can assign \pos tags to the \ybc and other text. There are two choices:

\noindent \textbf{(A) Convert the \ytb to the Yiddish script.}  An example would be going from the representation of the text in the tree (1) to the sentence (2). 
    This would need to be done for all the words in the two files from the \ytb. 
    
\noindent \textbf{(B) Convert the \ybc to \syo and then a romanized representation.}  An example would be  converting sentence (2), if it occurred in the \ybc, to the representation of the words in the tree (1).  This would need to be done for all 650 million words in the \ybc (and any text from other sources to be tagged and parsed in the future).

While the second alternative is maybe possible in principle, it would be very problematic to carry out.  It would need to be done automatically for the 650 million words, with their many alternate spellings and the inevitable \ocr errors.

In contrast, the first alternative requires the conversion of only a comparatively small amount of text from the \ytb, which to a certain extent can be checked manually. The first alternative also results in a tagger trained on standard Yiddish script, and therefore ready to use on new text in that form, without requiring further conversions.\footnote{Another possibility is to train a tokenizer to first split apart the words and then \pos tag words with modified tokenization, such as \<myyns .tw>.  We prefer to use a single model for the related tasks of tokenization and \pos tagging, and so we leave this possibility aside for now.}

\section{Converting \ytb to Yiddish Script}
\label{sec:conversion}

\begin{table}[t]
\begin{tabular}{|l|r|l|} \hline
tag & count & example word \\ \hline
 P\textasciitilde{}D        &  648 & oyf\textasciitilde{}n                \\ \hline
 RP\textasciitilde{}VBN     &  435 & on\textasciitilde{}gehoybn           \\ \hline
 RP\textasciitilde{}VB      &  249 & on\textasciitilde{}heybn             \\ \hline
 ADV\textasciitilde{}VBN    &  219 & avek\textasciitilde{}geshtelt        \\ \hline
 ADV\textasciitilde{}VB     &  140 & aroys\textasciitilde{}kumen          \\ \hline
 VBF\textasciitilde{}PRO    &  105 & bistu                \\ \hline
 MDF\textasciitilde{}PRO    &   76 & vestu                \\ \hline
 PRO\textasciitilde{}VBF    &   75 & kh'hob               \\ \hline
 PRO\textasciitilde{}HVF    &   59 & kh'hob               \\ \hline
 PRO\textasciitilde{}MDF    &   51 & kh'vel               \\ \hline
 NEG\textasciitilde{}ADV    &   49 & nit\textasciitilde{}o                \\ \hline
 P\textasciitilde{}WPRO     &   41 & far\textasciitilde{}vos              \\ \hline
 RP\textasciitilde{}TO\textasciitilde{}VB   &   33 & oyf\textasciitilde{}tsu\textasciitilde{}shteyn       \\ \hline
 WADV\textasciitilde{}Q     &   27 & vi\textasciitilde{}fl                \\ \hline
 RP\textasciitilde{}VAN     &   25 & on\textasciitilde{}geton             \\ \hline
 HVF\textasciitilde{}PRO    &   19 & hostu                \\ \hline
 VBI\textasciitilde{}PRO    &   19 & lo\textasciitilde{}mir               \\ \hline
 ES\textasciitilde{}VBF     &   18 & s'iz                 \\ \hline
 ADV\textasciitilde{}TO\textasciitilde{}VB  &   16 & avek\textasciitilde{}tsu\textasciitilde{}geyn        \\ \hline
 BEF\textasciitilde{}PRO    &   13 & bistu                \\ \hline
    \end{tabular}
    \caption{The 20 most common cases of combined \pos tags.   We also include the most common word for each combined tag.  In some cases we have inserted a \textasciitilde{} in the word as well, to indicate where it was originally split. \textit{kh'hob} appears twice since \textit{hob} can be HVF (auxiliary) or VBF (main verb). Likewise \textit{bistu} appears twice since \textit{bist} can be BEF (auxiliary) or VBF (main verb).}
    \label{tab:tags1}
\end{table}

The conversion of the two \ytb files just discussed to Yiddish script poses two problems, described in the following two subsections.

\subsection{Generating the original tokenization} 
\label{sec:conversion:transform}
As discussed in Section \ref{sec:ytb}, the tokens forming the leaves in the syntactic trees are sometimes modified forms of words from the original text.    The first aspect of the problem is to undo these modifications. 

\paragraph{Recombining words}
Section \ref{sec:ytb} showed an example of how some words were split apart for purposes of annotation.  We restored such cases based on information in the treebank about which words had been split. For example, in the tree (2),  {\it meyns} and {\it tu} are  combined as {\it meynstu}. %

Each word must have a single associated \pos label to be used for training/evaluation data for the \pos tagger, and so 
we connect the labels for each of the two split words with a \textasciitilde.  In the example at hand,
(VBF {\it meyns}) and (PRO {\it tu}) are combined as (VBF{\textasciitilde}PRO {\it meynstu}).
There are 53 such combined tags, which occur on 2,405 words. 
Table \ref{tab:tags1} shows the 20 most common ones,  accounting for 96.34\%  of the 2,405 words with combined tags.

\paragraph{Splitting words}
Sometimes, the inverse of the previous case occurs, in which the source text had two separate words that were combined as one in the treebank. An underscore in the word indicates words that were so combined, as in (WADV {\it vi\_azoy}), `how so' (literally `how so').  To generate the original tokenization for such cases,  we simply split apart the words at the underscore, and append a \_S0 and \_S1 to the label that is appropriate for the joined word. In this case, that results in the two tokens 
(WADV\_S0 {\it vi}) and (WADV\_S1 {\it azoy}). 

\begin{table}
\center
    \begin{tabular}{|l|r|l|} \hline
    tag & count & word \\ \hline
    ADV  & 39  & a\_mol \\ \hline
    Q    & 36  & a\_sakh \\ \hline
    NEG  & 25  & gor\_nit \\ \hline
    ADV  & 20  & mer\_nit \\ \hline
    ADV  & 17  & eyn\_mol \\ \hline
    WADV & 15  & vi\_azoy  \\ \hline
    \end{tabular}
    \caption{The six most common cases of words that were combined in the treebank and then split to generate the original tokenization.  They account for 64\% of all cases.
    \label{tab:tags2}}
\end{table}
There are 34 new tags, resulting from 236 originally separate words that were combined into one in the treebank.  The words that occur more than 10 times, accounting for 64\% of all such cases, are shown in Table \ref{tab:tags2}.

\paragraph{Results}
The processing in this section, with the combining and splitting of \ytb tokens, resulted in 82,675 tokens (compared to the 83,457 before).  The steps of recombining and splitting words resulted in a substantial increase in the total number of different \pos tags, from 68 to 155. Some of the combinations are very infrequent, and this is an area that will be revisited in future work. See Appendix \ref{app:tagset} for the complete tagset. 

\subsection{Conversion to Yiddish script}
At this point, with the creation of appropriate tokens, all that remains is to convert them to the original Yiddish script.  As discussed in Section \ref{sec:orthography}, the conversion from the romanized form to Yiddish script is not so straightforward, in significant part because there is no simple correspondence between the romanized form of a ``non-phonetic'' component word and its representation in Yiddish script.   Another example of this in (1), with the romanized name {\it rokhl} corresponding to \<r.hl>.  Applying the rules for the ``phonetic'' component to {\it rokhl} would result in the incorrect 
 \<r'Akl>, which has an extra vowel and uses the 
 \<k|> character instead of \<.h> for {\it kh}.  

We carried out the conversion with a wrapper around calls to the \texttt{yiddish} Python library.\footnote{\url{https://github.com/ibleaman/yiddish}}  The library includes code for converting between the different representations, utilizing an extensive list of such non-phonetic cases.  The wrapper overrides the results in some cases.  For example, in some cases we used the \pos tag of a word to determine the correct conversion, such as for the \textit{shem} case mentioned in Section \ref{sec:orthography}.  This is discussed, along with the some more details of our usage of the \texttt{yiddish} package, in Appendix \ref{app:conversion}.\footnote{The software for processing the \ytb as discussed in this section is at 
\url{https://github.com/skulick/ppchyprep}. \label{fn:ppchyprep}}

\section{Embeddings Trained on the YBC}
\label{sec:embeddings}

\subsection{Contextualized embeddings}
\label{sec:embeddings:con}
As mentioned in Section \ref{sec:previous}, the \xlmrobertabase 
multilingual embeddings include a (relatively) small amount of Yiddish in 
their training data.  We carry out continued pretraining using \ybc, for 1, 2, 5, and 10 epochs. For more details of the training of the embeddings, see Appendix \ref{app:con_embeddings}.

\begin{table}[t]
    \centering
    \begin{tabular}{|l|r|c|r|} \hline
     \multicolumn{2}{|c|}{word}         & similarity  & count \\ \hline 
     \multicolumn{4}{|c|}{Example 1}  \\ \hline
      {\it badaytung}  & \<b'adyaya.twng>  & -  &     788 \\ \hline
      {\it badeytung}  & \<b'adyy.twng> & 0.65 &   12,082 \\ \hline
      {\it baditung}   & \<b'ady.twng> & 0.58 &     141 \\ \hline
      \multicolumn{4}{|c|}{Example 2}  \\ \hline
      {\it ehnlikh}    & \<`hnlyK> &  - &    6,521 \\ \hline
      {\it enlekh}     & \<`nl`K> & 0.65 &   18,485 \\ \hline
      {\it enlikh}     & \<`nlyK> & 0.61 &    1,822 \\ \hline
       \multicolumn{4}{|c|}{Example 3}  \\ \hline
      {\it ems'n}      &  \<'mt\textrm{'}n> & - & 12,743 \\ \hline
      {\it ems'en}     &  \<'mt\textrm{'}`n> & 0.75 & 4,191 \\ \hline
      {\it emsn}       &  \<'mtn> & 0.72 & 27,129 \\ \hline
      \multicolumn{4}{|c|}{Example 4}  \\ \hline
      {\it bur}        & \<bwr> & -  &    6,876  \\ \hline
      {\it bud}        & \<bwd> & 0.73 &    9,615  \\ \hline
      {\it bukh}       &  \<bwK> & 0.72 &   97,877  \\ \hline
    \end{tabular}
    \caption{Sample similarities with word embeddings trained on the \ybc corpus. For each of the four examples, the second and third words listed are the closest to the first word by cosine similarity with the \glove embeddings. The last column is the number of occurrences of that word in the \ybc corpus.}
    \label{tab:embeddings}
\end{table}

\subsection{Non-contextualized embeddings}
\label{sec:embeddings:noncon}
Non-contextualized embeddings such as \glove \cite{pennington2014glove}, where every occurrence of a word has a single vector, perform worse than
contextualized embeddings such as \xlmrobertabase,
where each occurrence of a word may have a different vector.  However, non-contextualized embeddings can still be useful to illustrate the relations among embeddings.  Here we use \glove embeddings trained on the \ybc corpus, with a dimension of 300. See Appendix \ref{app:noncon_embeddings} for further details on the training of these embeddings.  

Table \ref{tab:embeddings} shows four examples of the embedding relationships.  For each of the examples, we have selected one word and identified the two  most ``similar'' words by finding the words with the highest cosine similarity to them based on the \glove embeddings.  The similarity is a score between 0.0 and 1.0, where 1.0 means perfect distributional similarity in the \ybc corpus.  

\subsubsection{Embeddings identifying alternate spellings}
\label{sec:embeddings:alt}

The first word listed is {\it badaytung}  `significance,  meaning'. The two closest words to it are alternate spellings.  In the first, {\it badeytung}, 
{\it ay} (\<yaya>) is written as {\it ey} (\<yy>), without the diacritic marker below, as is often the case.  In fact, while {\it badaytung} is the correct 
 spelling, the version without the diacritic is far more common (12,082 vs.~788).  The next word, {\it baditung}, is another nonstandard spelling, though a very infrequent one (141). 

This example shows how the embeddings can capture relationships among alternate spellings without requiring prior ``standardization'' of the texts.
The second example is another such case.  The word is the nonstandard
spelling for {\it ehnlikh} `similar'. The word closest to it by cosine similarity is {\it enlekh}, which is actually the ``standard'' form, and in this case it is far more frequent than the nonstandard form, which differs both in the vowel ({\it i} for {\it e}), and in the extra {\it h}, added by analogy to the spelling of the cognate word in German.
The next word, {\it enlikh}, is yet another variant spelling.

The third example returns to the example mentioned in Section \ref{sec:ybc}.  The two variants, {\it ems'n} and {\it emsn}, are in a close relationship, as we hoped would be the case. In addition, our procedures identifies yet another variant, {\it ems'en}, with an extra {\it e} before the final {\it n}.\footnote{We have limited ourselves in these examples to the first two most similar words.  If the list is expanded, our procedure also finds synonyms and so on, as is typical for such embeddings.}$^{,}$\footnote{There are many other cases of orthographic variation to consider, such as 
inconsistent orthographic variation with separate whitespace-delimited tokens.  Future work with contextualized embeddings will consider such cases in the context of the \pos-tagging and parsing accuracy.}

\subsubsection{Embeddings identifying \ocr errors}
\label{sec:embeddings:ocr}
The fourth example is somewhat different.  Here the word {\it bur} `ignoramus, boor' is an actual word, as is the word closest to it, {\it bud} `shop, stall'.  However, given their distributional similarity and the similarity in the final letters, which could be easily confused, what seems far more likely is that these are \ocr errors for the next word closest to them, {\it bukh} `book', and this is indeed the case for the instances of {\it bur} and {\it bud} 
that we have spot-checked. 

This example shows the potential for word embeddings trained on the \ybc collection to help find \ocr errors for correction.  Ideally, integrating contextualized embeddings into an \ocr detection system could find such cases on a sentence-by-sentence basis - that is, deciding for each occurrence of {\it bur} or {\it bud} if it is a likely \ocr error in that sentence or a genuine instance of the word \cite{postocr,icdar2019}.

\section{\pos Tagger Training and Evaluation}
\label{sec:model}
We trained a tagger using five versions of the 
\xlmrobertabase embeddings - the standard version and four variants with different amounts of continued pretraining on \ybc, as discussed in Section \ref{sec:embeddings:con}. For completeness, we also also trained the tagger using the \glove embeddings discussed in Section \ref{sec:embeddings:noncon}.

Our POS tagger is implemented using the \flair library \cite{akbik2019flair}, slightly adapting their sample script for training.   The \xlmrobertabase embeddings are used in a model consisting of a linear layer on top of the transformer embeddings, predicting a score for each \pos tag.   Additionally, we incorporate parameters for the transitions between each pair of \pos tags so that the resulting model has the form of a linear chain conditional random field (CRF). Training is conducted using a sentence-level CRF loss computed using forward-backward, while Viterbi decoding is used for inference at test time.  The model using the \glove embeddings is similar, with the addition of another linear layer and bidirectional LSTM on top of the embeddings before the linear layer and CRF. See Appendix
\ref{app:model} for further details.

The amount of training and evaluation data we have (82,675 tokens) is very small compared e.g. to \pos taggers trained on the  1 million words of the \ptb.  With such a small amount of data for training and evaluation, and that from only two sources, we used a 10-fold stratified split.  Each of the 10 splits has 90\% of the text for training, 5\% for validation, and 5\% for testing.  The validation section is used to select the best model during training.  The training, validation, and test sections are all balanced between the two sources, with 79\% from \citet{olsvanger} and 21\% from \citet{grinefelder}.

\section{Results}
\label{sec:results}

\begin{table}[t]
    \centering
    \begin{tabular}{|l|c|c|c|} \hline
    & \multicolumn{2}{|c|}{accuracy} \\
    \cline{2-3}
    embeddings & validation & test \\ \hline
     glove     & 96.71 (0.39) & 96.35 (0.25) \\ \hline
 xlm-rb (0) & 98.06 (0.27) & 97.71 (0.29) \\ \hline
 xlm-rb (1)  & 98.40 (0.28) & 98.23 (0.25) \\ \hline
 xlm-rb (2) & 98.46 (0.26) & 98.26 (0.23) \\ \hline
 xlm-rb (5) & 98.50 (0.22) & 98.21 (0.23) \\ \hline
 xlm-rb (10) & 98.44 (0.20) & 98.21 (0.26) \\ \hline
    \end{tabular}
    \caption{Cross-validated \pos tagging results, for the six configurations of the \pos tagger, differing on the embeddings used. The first row shows the results with \glove embeddings, and the other rows show the results using the \xlmrobertabase with the number of epochs of continued pretraining on \ybc in parentheses.  The row 
    \texttt{xlm-rb~(0)} is \xlmrobertabase without any continued pretraining.  Each cell shows the mean and standard deviation of the \pos accuracy over the 10 validation or test sections.}
    \label{tab:posresults}
\end{table}

\begin{table}[t]
{\small
 \begin{tabular}{|l|r|c|c|} \hline
  &  & \multicolumn{2}{|c|}{F1} \\
  \cline{3-4}
            & count\ \         & xlm-rb (0)     & xlm-rb (2) \\ \hline \hline
 Total      & 4145.40  & 97.71 (00.29)  & 98.26 (00.23)  \\ \hline
 no PUNC & 3391.50  & 97.21 (00.34)  & 97.87 (00.28)  \\ \hline
 tilde  & 123.40   & 94.76 (01.98)  & 96.20 (01.49)  \\ \hline \hline
 PUNC       & 753.90  & 99.99 (00.02)  & 99.99 (00.02)  \\ \hline
 N          & 523.90  & 97.45 (00.53)  & 98.19 (00.50)  \\ \hline
 PRO        & 398.50  & 99.30 (00.20)  & 99.49 (00.29)  \\ \hline
 D          & 333.50  & 99.64 (00.29)  & 99.62 (00.30)  \\ \hline
 VBF        & 244.60   & 97.79 (00.65)  & 98.50 (00.51)  \\ \hline
 ADV        & 205.80  & 95.43 (01.17)  & 96.68 (00.97)  \\ \hline
 P          & 193.60  & 98.06 (00.83)  & 98.56 (00.72)  \\ \hline
 NPR        & 136.80  & 98.44 (00.55)  & 98.62 (00.62)  \\ \hline
 VBN        & 132.00  & 98.44 (00.68)  & 98.84 (00.73)  \\ \hline
 CONJ       & 123.70  & 99.30 (00.62)  & 99.20 (00.74)  \\ \hline
 VB         & 119.00  & 97.64 (00.98)  & 98.58 (00.68)  \\ \hline
 HVF        & 104.60   & 99.85 (00.24)  & 99.96 (00.14)  \\ \hline
 ADJ        & 100.40   & 90.92 (02.66)  & 94.11 (01.87)  \\ \hline
 MDF        & 84.20   & 98.29 (01.23)  & 98.35 (01.32)  \\ \hline
 C          & 70.80   & 96.45 (01.55)  & 97.07 (01.68)  \\ \hline
 NEG        & 57.80    & 99.20 (00.71)  & 99.47 (00.58)  \\ \hline
 BEF        & 51.20   & 98.77 (00.93)  & 98.73 (00.99)  \\ \hline
 Q          & 45.90   & 96.73 (01.76)  & 97.53 (01.27)  \\ \hline
 WPRO       & 43.30  & 96.21 (02.14)  & 97.34 (01.06)  \\ \hline
 INTJ       & 35.90    & 93.52 (02.89)  & 94.85 (03.31)  \\ \hline
 P\textasciitilde{}D        & 35.50    & 97.87 (01.54)  & 98.67 (01.53)  \\ \hline
 H          & 24.70    & 84.46 (10.13)  & 85.45 (09.24)  \\ \hline
 PRO\$       & 24.10    & 98.95 (01.52)  & 100.00 (00.00) \\ \hline
 VBI        & 23.00    & 89.69 (05.16)  & 93.11 (05.85)  \\ \hline
 RP\textasciitilde{}VBN     & 21.80   & 97.80 (02.78)  & 98.18 (01.83)  \\ \hline
\end{tabular}}
\caption{A breakdown of the accuracy scores for the test section for the xlm-rb~(0) and xlm-rb~(2) configurations from Table \ref{tab:posresults}, showing the mean and standard deviation F1 scores for each tag over the 10 test sections.   The ``count'' column reports the mean number of occurrences of the \pos tag over the 10 test sections.}
\label{tab:posresults2}
\end{table}

\subsection{Results on the \ytb evaluation}
Table \ref{tab:posresults} shows the results for each of the six configurations and for the validation and test sections. Each cell shows the mean and standard deviation across the 10 splits. Unsurprisingly, the results using \glove are below the results using \xlmrobertabase.  The one epoch of continued pretraining further improves the results, and further pretraining results in incremental increases and then a dropoff. 

To get a better idea of the performance and improvement on particular tags, in Table \ref{tab:posresults2} we break down the (xlm-rb(0), `test') and (xlm-rb(2), `test) cells from Table \ref{tab:posresults} into F1 scores.  The first line has the totals, just as in Table \ref{tab:posresults}.\footnote{The ``Total'' F1 score for all tags is equivalent to the accuracy.} 
Since PUNC, the most common tag, has an almost perfect score, the second line shows the score for all tags excluding PUNC.  We were also particularly interested in how well the tagger does on the complex tags with a tilde separating the components (as discussed in Section \ref{sec:conversion:transform}), and this result is shown in the third row.\footnote{These tags will be important for a next, parsing, phase of the work, since they determine the tokenization needed for the syntactic trees.}  While the total w/o PUNC increases by 0.66 (97.21 to 97.87) with two epochs of pretraining,  the combined tags increase by 1.44 (from 94.76 to 96.20). Below those three lines, we show the scores for the individual tags.  They show a general increase, with a few exceptions, across all tags with the two epochs of pretraining.

\subsection{Limitations of the evaluation}
\label{sec:results:limitations}
The evaluation just presented is significantly incomplete because of the limited amount and range of gold-standard annotated data.  The test sections all come from the \ytb corpus.  As discussed above, we aim to tag and parse the \ybc corpus, and possibly beyond that.  Given its small size, the \ytb text has a necessarily limited vocabulary.  Moreover, it does not exhibit  the spelling variations seen in  the \ybc corpus.  Since the embeddings trained on the \ybc should allow the model to further  generalize beyond the \ytb training data, we expect to see a significant further divergence between the scores when evaluating on  text from the \ybc.  This would likely be even more salient for a syntactic parser.

Having some gold-annotated \pos text from the \ybc corpus is therefore a significant need, and preferably with syntactic annotation as well, in preparation for next steps on this work, when we expand from \pos tagging to syntactic parsing.

\section{Conclusion and Next Steps}
\label{sec:conclusion}

We have presented here the first steps toward the larger goal of training a \pos tagger and syntactic parser to automatically annotate a large corpus of Yiddish.  We trained embeddings on the \ybc corpus and created a unified representation for the annotated data from the \ytb and the unannotated \ybc.  Using the framework based on a cross-validation split for training and evaluating a \pos tagger trained on the \ytb, we showed that even with such limited training and evaluation data, continued pretraining on embeddings using \ybc resulted in improved performance. 

\subsection{Future work}
As discussed in Section \ref{sec:results:limitations}, we require more gold \pos-annotated text to evaluate the tagger on. We plan to create the requisite data by tagging samples from the \ybc corpus and manually correcting the predicted \pos tags.

We will also be training and evaluating a syntactic parser in addition to the \pos tagger, using the \ytb annotation.  This will mostly follow the training/evaluation framework established above.  Just as with the \pos tagger, we will need additional evaluation data, this time manually annotated with gold syntactic trees.  Given the greater complexity of a parser, the need for gold data is correspondingly more pressing. 

We are also interested in \pos-tagging the \ybc corpus and in exploring the possibility mentioned at the end of the introduction and in Section \ref{sec:embeddings:ocr}
of using the methodology and results reported here to augment the search capabilities on the \ybc, to identify variant spellings, and to locate \ocr errors for correction.

\section{Acknowledgments}
We would like to thank the Yiddish Book Center for making available the \ocr'd texts of the book collection.  The work described
in this paper would otherwise of course have not been possible.

\bibliography{ppcbib}

\begin{thebibliography}{43}
\expandafter\ifx\csname natexlab\endcsname\relax\def\natexlab#1{#1}\fi

\bibitem[{Akbik et~al.(2019)Akbik, Bergmann, Blythe, Rasul, Schweter, and
  Vollgraf}]{akbik2019flair}
Alan Akbik, Tanja Bergmann, Duncan Blythe, Kashif Rasul, Stefan Schweter, and
  Roland Vollgraf. 2019.
\newblock {FLAIR}: An easy-to-use framework for state-of-the-art {NLP}.
\newblock In \emph{{NAACL} 2019, 2019 Annual Conference of the North American
  Chapter of the Association for Computational Linguistics (Demonstrations)},
  pages 54--59.

\bibitem[{Blum(2015)}]{blum2015}
Yakov~Peretz Blum. 2015.
\newblock \href {https://academicworks.cuny.edu/gc_etds/525} {Techniques for
  automatic normalization of orthographically variant {Y}iddish texts}.
\newblock Master's thesis, CUNY.

\bibitem[{Conneau et~al.(2020)Conneau, Khandelwal, Goyal, Chaudhary, Wenzek,
  Guzm{\'a}n, Grave, Ott, Zettlemoyer, and
  Stoyanov}]{conneau-etal-2020-unsupervised}
Alexis Conneau, Kartikay Khandelwal, Naman Goyal, Vishrav Chaudhary, Guillaume
  Wenzek, Francisco Guzm{\'a}n, Edouard Grave, Myle Ott, Luke Zettlemoyer, and
  Veselin Stoyanov. 2020.
\newblock \href {https://doi.org/10.18653/v1/2020.acl-main.747} {Unsupervised
  cross-lingual representation learning at scale}.
\newblock In \emph{Proceedings of the 58th Annual Meeting of the Association
  for Computational Linguistics}, pages 8440--8451, online. Association for
  Computational Linguistics.

\bibitem[{Ecay(2015)}]{ecay}
Aaron Ecay. 2015.
\newblock \href {https://repository.upenn.edu/edissertations/1049/} {\emph{A
  multi-step analysis of the evolution of {E}ng\-lish do-support}}.
\newblock Ph.D. thesis, University of Pennsylvania.

\bibitem[{Galves(2020)}]{galves-20}
Charlotte Galves. 2020.
\newblock Relaxed {V-Second} in {C}lassical {P}ortuguese.
\newblock In Rebecca Woods and Sam Wolfe, editors, \emph{Rethinking
  {Verb-Second}}, pages 368--395. Oxford University Press.

\bibitem[{Galves et~al.(2017)Galves, Andrade, and Faria}]{tycho-brahe}
Charlotte Galves, Aroldo Leal~de Andrade, and Pablo Faria. 2017.
\newblock \href {http://www.tycho.iel.unicamp.br/~tycho/corpus/texts/psd.zip}
  {Tycho {B}rahe {P}arsed {C}orpus of {H}istorical {P}ortuguese}.

\bibitem[{Gold(1977)}]{gold}
David~L. Gold. 1977.
\newblock Successes and failures in the standardization and implementation of
  {Y}iddish spelling and romanization.
\newblock In Joshua~A. Fishman, editor, \emph{Advances in the Creation and
  Revision of Writing Systems}. De Gruyter Mouton.

\bibitem[{Grave et~al.(2018)Grave, Bojanowski, Gupta, Joulin, and
  Mikolov}]{grave-etal-2018-learning}
Edouard Grave, Piotr Bojanowski, Prakhar Gupta, Armand Joulin, and Tomas
  Mikolov. 2018.
\newblock \href {https://aclanthology.org/L18-1550} {Learning word vectors for
  157 languages}.
\newblock In \emph{Proceedings of the Eleventh International Conference on
  Language Resources and Evaluation ({LREC} 2018)}, Miyazaki, Japan. European
  Language Resources Association (ELRA).

\bibitem[{Hirshbein(1951)}]{grinefelder2}
Peretz Hirshbein. 1951.
\newblock \href
  {https://www.yiddishbookcenter.org/collections/yiddish-books/spb-nybc207252/hirschbein-peretz-grine-felder-trilogye}
  {\emph{Grine Felder}}.
\newblock Aber Press, Inc.

\bibitem[{Hirshbein(1977)}]{grinefelder}
Peretz Hirshbein. 1977.
\newblock Grine felder [{G}reen fields].
\newblock In Hyman Bass, editor, \emph{Di yidishe drame fun 20stn yorhundert
  [20th-century {Y}iddish drama]}, pages 61--106. Congress for Jewish Culture.

\bibitem[{Jacobs(2005)}]{jacobs}
Neil~G. Jacobs. 2005.
\newblock \emph{Yiddish: A linguistic introduction}.
\newblock Cambridge University Press, New York.

\bibitem[{Kahn(2017)}]{kahn}
Lily Kahn. 2017.
\newblock Yiddish.
\newblock In Lily Kahn and Aaron~D. Rubin, editors, \emph{Handbook of Jewish
  Language: Revised and Updated Edition}. Brill.

\bibitem[{Kirjanov et~al.(2014)Kirjanov, Orehov, and Panova}]{kirjanov}
D.P. Kirjanov, B.V Orehov, and T.A. Panova. 2014.
\newblock \href {https://publications.hse.ru/en/articles/139396314} {Yiddish
  orthographies variety and problems of automatic transliteration}.

\bibitem[{Kroch(2020)}]{ppche}
Anthony Kroch. 2020.
\newblock \href {https://doi.org/doi.org/10.35111/4hzx-5483} {Penn {P}arsed
  {C}orpora of {H}istorical {E}nglish}.
\newblock LDC2020T16 Web Download. Philadelphia: Linguistic Data Consortium.
\newblock Contains {P}enn-{H}elsinki {P}arsed {C}orpus of {M}iddle {E}nglish,
  second edition, {P}enn-{H}elsinki {P}arsed {C}orpus of {E}arly {M}odern
  {E}nglish, and {P}enn {P}arsed {C}orpus of {M}odern {B}ritish {E}nglish.

\bibitem[{Kroch and Santorini(2021)}]{ppchf}
Anthony Kroch and Beatrice Santorini. 2021.
\newblock \href {https://github.com/beatrice57/mcvf-plus-ppchf} {{Penn-{BFM}
  {P}arsed {C}orpus of {H}istorical {F}rench}, version 1.0}.

\bibitem[{Kroch et~al.(2016)Kroch, Santorini, and Diertani}]{ppcmbe}
Anthony Kroch, Beatrice Santorini, and Ariel Diertani. 2016.
\newblock \href
  {http://www.ling.upenn.edu/ppche/ppche-release-2016/PPCMBE2-RELEASE-1}
  {{P}enn {P}arsed {C}orpus of {M}odern {B}ritish {E}nglish ({PPCMBE2})}.
\newblock CD-ROM, second edition, release 1.

\bibitem[{Kroch et~al.(2000{\natexlab{a}})Kroch, Taylor, and
  Ringe}]{kroch-taylor-ringe}
Anthony Kroch, Ann Taylor, and Donald Ringe. 2000{\natexlab{a}}.
\newblock The {M}iddle {E}nglish verb-second constraint: A case study in
  language contact and language change.
\newblock In Susan Herring, Lene Schoessler, and Peter van Reenen, editors,
  \emph{Textual parameters in older language}, pages 353--391. Benjamins.

\bibitem[{Kroch et~al.(2000{\natexlab{b}})Kroch, Taylor, and
  Santorini}]{ppcme2}
Anthony Kroch, Ann Taylor, and Beatrice Santorini. 2000{\natexlab{b}}.
\newblock \href
  {http://www.ling.upenn.edu/ppche/ppche-release-2016/PPCME2-RELEASE-4}
  {{P}enn-{H}elsinki {P}arsed {C}orpus of {M}iddle {E}nglish (\ppcme)}.
\newblock CD-ROM, second edition, release 4.

\bibitem[{Kulick et~al.(2022{\natexlab{a}})Kulick, Ryant, and
  Santorini}]{kulick-etal-2022-penn}
Seth Kulick, Neville Ryant, and Beatrice Santorini. 2022{\natexlab{a}}.
\newblock \href {https://doi.org/10.18653/v1/2022.findings-naacl.44} {\
  {P}enn-{H}elsinki parsed corpus of early {M}odern {E}nglish: First parsing
  results and analysis}.
\newblock In \emph{Findings of the Association for Computational Linguistics:
  NAACL 2022}, pages 578--593. Association for Computational Linguistics.

\bibitem[{Kulick et~al.(2022{\natexlab{b}})Kulick, Ryant, and
  Santorini}]{kulick-etal-2022-parsing}
Seth Kulick, Neville Ryant, and Beatrice Santorini. 2022{\natexlab{b}}.
\newblock \href {https://aclanthology.org/2022.scil-1.12} {{P}arsing {E}arly
  {M}odern {E}nglish for linguistic search}.
\newblock In \emph{Proceedings of the Society for Computation in Linguistics
  2022}. Association for Computational Linguistics.

\bibitem[{Kulick et~al.(2023)Kulick, Ryant, and
  Santorini}]{kulick-etal-2023-parsing}
Seth Kulick, Neville Ryant, and Beatrice Santorini. 2023.
\newblock \href {https://scholarworks.umass.edu/scil/vol6/iss1/21/} {Parsing
  ``{E}arly {E}nglish {B}ooks {Online}'' for linguistic search}.
\newblock In \emph{Proceedings of the Society for Computation in Linguistics
  2023}, online. Proceeding of the Society for Computation in Linguistics.

\bibitem[{Kutzik(2019)}]{kutzik}
Jordan Kutzik. 2019.
\newblock \href
  {https://forward.com/yiddish/435210/10-000-yiddish-books-now-fully-searchable-online/}
  {10,000 {Y}iddish books now fully searchable online}.
\newblock \emph{The Forward}.
\newblock 11/20/2019.

\bibitem[{Marcus et~al.(1993)Marcus, Santorini, and
  Marcinkiewicz}]{marcus-etal-1993-building}
Mitchell~P. Marcus, Beatrice Santorini, and Mary~Ann Marcinkiewicz. 1993.
\newblock \href {https://www.aclweb.org/anthology/J93-2004} {Building a large
  annotated corpus of {E}nglish: The {P}enn {T}reebank}.
\newblock \emph{Computational Linguistics}, 19(2):313--330.

\bibitem[{Markey(2012)}]{Markey}
Keven Markey. 2012.
\newblock \href
  {https://www.yiddishbookcenter.org/language-literature-culture/pakn-treger/assaf-urieli-computational-yiddish-linguist}
  {Assaf {U}rieli: Computational {Y}iddish linguist}.
\newblock \emph{Pakn Treger}, Fall(66).

\bibitem[{Martineau et~al.(2021)Martineau, Hirschbühler, Kroch, and
  Morin}]{mcvf}
France Martineau, Paul Hirschbühler, Anthony Kroch, and Yves~Charles Morin.
  2021.
\newblock \href {https://github.com/beatrice57/mcvf-plus-ppchf} {{{MCVF}
  {C}orpus}, parsed, version 2.0}.

\bibitem[{Nguyen et~al.(2021)Nguyen, Jatowt, Coustaty, and Doucet}]{postocr}
Thi Tuyet~Hai Nguyen, Adam Jatowt, Mickael Coustaty, and Antoine Doucet. 2021.
\newblock \href {https://doi.org/10.1145/3453476} {Survey of post-{OCR}
  processing approaches}.
\newblock \emph{ACM Comput. Surv.}, 54(6).

\bibitem[{Olsvanger(2022)}]{olsvanger2022}
Emanuel Olsvanger. 2022.
\newblock \emph{Raisins and Almonds - Jokes, Tales, Legends}.
\newblock Leyvik Publishing House.
\newblock Prepared for publication by Simon Noyberg.

\bibitem[{Olsvanger(1947)}]{olsvanger}
Immanuel Olsvanger. 1947.
\newblock \emph{Röyte pomerantsen, Jewish folk humor}.
\newblock Schocken Books.

\bibitem[{Pennington et~al.(2014)Pennington, Socher, and
  Manning}]{pennington2014glove}
Jeffrey Pennington, Richard Socher, and Christopher~D. Manning. 2014.
\newblock \href {http://www.aclweb.org/anthology/D14-1162} {Glove: Global
  vectors for word representation}.
\newblock In \emph{Empirical Methods in Natural Language Processing (EMNLP)},
  pages 1532--1543.

\bibitem[{Prince(1993)}]{prince1993discourse}
Ellen~F. Prince. 1993.
\newblock On the discourse functions of syntactic form in {Y}iddish: Expletive
  {ES} and subject postposing.
\newblock In \emph{The Field of {Y}iddish. 5th collection. Studies in {Yiddish}
  language, folklore, and literature.} Northwestern University Press and YIVO
  Institute for Jewish Research.

\bibitem[{Rigaud et~al.(2019)Rigaud, Doucet, Coustaty, and Moreux}]{icdar2019}
Christophe Rigaud, Antoine Doucet, Mickaël Coustaty, and Jean-Philippe Moreux.
  2019.
\newblock \href {https://doi.org/10.1109/ICDAR.2019.00255} {{ICDAR} 2019
  competition on post-{OCR} text correction}.
\newblock In \emph{2019 International Conference on Document Analysis and
  Recognition (ICDAR)}, pages 1588--1593.

\bibitem[{Saleva(2020)}]{saleva-2020-multi}
Jonne Saleva. 2020.
\newblock \href {https://aclanthology.org/2020.lrec-1.119} {A multi-orthography
  parallel corpus of {Y}iddish nouns}.
\newblock In \emph{Proceedings of the 12th Language Resources and Evaluation
  Conference}, pages 948--952, Marseille, France. European Language Resources
  Association.

\bibitem[{Santorini(1989)}]{santorini1989diss}
Beatrice Santorini. 1989.
\newblock \emph{The generalization of the verb-second constraint in the history
  of {Y}iddish}.
\newblock Ph.D. thesis, University of Pennsylvania.

\bibitem[{Santorini(1992)}]{santorini1992variation}
Beatrice Santorini. 1992.
\newblock \href
  {https://www.semanticscholar.org/paper/Variation-and-change-in-Yiddish-subordinate-clause-Santorini/be757433424465e5a6f32d92fd2421bc7a3cefe6}
  {Variation and change in {Y}iddish subordinate clause word order}.
\newblock \emph{Natural Language and Linguistic Theory}, 10(4):595--640.

\bibitem[{Santorini(1993)}]{santorini1993rate}
Beatrice Santorini. 1993.
\newblock The rate of phrase structure change in the history of {Y}iddish.
\newblock \emph{Language Variation and Change}, 5(3):257--283.

\bibitem[{Santorini(2021)}]{yiddishtb}
Beatrice Santorini. 2021.
\newblock \href
  {https://github.com/beatrice57/penn-parsed-corpus-of-historical-yiddish}
  {Penn {P}arsed {C}orpus of {H}istorical {Y}iddish, v1.0}.

\bibitem[{Taylor et~al.(2006)Taylor, Nurmi, Warner, Pintzuk, and
  Nevalainen}]{pceec}
Ann Taylor, Arja Nurmi, Anthony Warner, Susan Pintzuk, and Terttu Nevalainen.
  2006.
\newblock \href {http://www-users.york.ac.uk/~lang22/PCCEC-manual/index.htm}
  {Parsed {C}orpus of {E}arly {E}ng\-lish {C}orrespondence}.
\newblock Distributed by the {O}xford {T}ext {A}rchive.

\bibitem[{Taylor et~al.(2003)Taylor, Warner, Pintzuk, and Beths}]{ycoe}
Ann Taylor, Anthony Warner, Susan Pintzuk, and Frank Beths. 2003.
\newblock {York-{T}oronto-{H}elsinki {P}arsed {C}orpus of {O}ld {E}ng\-lish
  {P}rose}.
\newblock Distributed by the {O}xford {T}ext {A}rchive.

\bibitem[{Urieli and Vergez-Couret(2013)}]{urieli2013}
Assaf Urieli and Marianne Vergez-Couret. 2013.
\newblock \href {https://dumas.ccsd.cnrs.fr/AO-LINGUISTIQUE/hal-00979665v1}
  {Jochre, océrisation par apprentissage automatique : Étude comparée sur le
  yiddish et l'occitan}.
\newblock In \emph{TALARE 2013 : Traitement automatique des langues régionales
  de France et d'Europe}. Les Sables d'Olonne, France.

\bibitem[{Wallenberg(2016)}]{wallenberg-2016}
Joel~C. Wallenberg. 2016.
\newblock Extraposition is disappearing.
\newblock \emph{Language}, 92(4):e237--e256.

\bibitem[{Wallenberg et~al.(2021)Wallenberg, Bailes, Cuskley, and
  Ingason}]{wallenberg-et-al-2021}
Joel~C. Wallenberg, Rachael Bailes, Christine Cuskley, and Anton~Karl Ingason.
  2021.
\newblock \href {https://doi.org/10.3390/languages6020060} {Smooth signals and
  syntactic change}.
\newblock \emph{Languages}, 6(2):60.

\bibitem[{Wallenberg et~al.(2011)Wallenberg, Ingason, Sigurðsson, and
  Rögnvaldsson}]{icepahc}
Joel~C. Wallenberg, Anton~Karl Ingason, Einar~Freyr Sigurðsson, and Eiríkur
  Rögnvaldsson. 2011.
\newblock \href {http://www.linguist.is/icelandic_treebank} {{Icelandic
  {P}arsed {H}istorical {C}orpus (IcePaHC)}, v0.9}.

\bibitem[{Weinreich(2011)}]{weinreich}
Uriel Weinreich. 2011.
\newblock \emph{College Yiddish (Sixth Revised Edition)}.
\newblock YIVO Institute for Jewish Research.

\end{thebibliography}

\appendix

\section{The Yiddish Book Center Corpus}
\label{app:ybc}
\subsection{Initial downloading and text extraction}
We first carried out the following steps:
\begin{enumerate}[leftmargin=*]
    \item We downloaded 10,970 MARC files from the {\tt nationalyiddishbookcenter} collection on the internet archive.
    \item For each such MARC file {\tt <id>}, we downloaded the corresponding \html file at  \\
    {\small \tt https://ocr.yiddishbookcenter.org/}
    {\small \tt contents?doc=<id>}.
    \item 9,925 of the MARC files had a corresponding \html file with the OCR text.\footnote{The html files were downloaded on April 9, 2021.}
    \item We extracted the \ocr content from each file, removing the header and footer and so on. We retained the page number indicators.  
    \item We normalized all the files with NFC normalization, and also did some additional minor normalization, such as changing occurrences of ``no-break space'' (unicode a0) to a regular space, and the em dash (2014) to a  hyphen. 
    \item We removed 120 files containing very rare characters, leaving 9,805 files to work with.
\end{enumerate}

While extracting the OCR content, we also stored each line as two tab-delimited fields, where the second is the actual text, and the first is a sentence identifier with the page number of the source pdf (extracted from the page number attribute in the html), together with the line number within that page.  This proved useful for looking up particular cases of words. 

\section{Aspects of the \ybc}
\label{app:aspects}
\subsection{Apostrophes as commas} 
It is sometimes the case that an apostrophe is \ocr'd as a comma - e.g. in one instance \<k|\textrm{'}.tr'Ag> {\it kh'trog} `I carry' is \ocr'd as \<k|\textrm{,}.tr'Ag> {\it kh,trog}, which then gets tokenized as \mbox{\<k|\textrm{ , }.tr'Ag>} {\it kh , trog}, resulting in three separate tokens.  In future processing we will adjust the tokenization to account for such cases.  
\subsection{Pointed Yiddish characters}
\label{app:aspects:pointed}
There are some files in which the Yiddish script has extensive use of diacritics  to (redundantly) represent vowels, lowering the \ocr accuracy for such files.  
For example, an instance of  \<gE`zA'g:.t> is \ocr'd as \<g`N'q.t>.  
In future processing we will omit these files. We are not currently sure of how many files this affects, but it appears to be relatively few.

\subsection{Unusual words and \ocr errors}
\label{app:aspects:unusual}
We sorted all the word tokens by frequency (after the tokenization step described in Section \ref{sec:ybc}) and examined some cases that seemed unexpected.  There were two types of cases we looked at, as described below in Sections \ref{app:aspects:unusual:1} and \ref{app:aspects:unusual:2}.  We made no further study of the unexpected character sequences beyond the cursory checking described. Our goal was to start exploring how combining this sort of analysis 
with cases such as the fourth example in Section \ref{sec:embeddings} and some combination of manual and automated correction might be used for post-\ocr text correction. 

\subsubsection{Words with a word-medial final form}
\label{app:aspects:unusual:1}
There are five characters in the Yiddish script with a form that only appears at the end of a word (or before a hyphen in a compound).  A search of the word list for exceptions to this rule revealed 269,814 different words, 2.5\% of the 10,642,884 different words in the corpus.  They totalled 427,855 occurrences, 0.07\% of the 653,326,190 words in the corpus.  237,926 of the 269,814 words (88.2\%) occur only once in the corpus.  

While such cases are relatively infrequent, we were curious as to what they could be.  Some of the most common cases were not \ocr errors, but rather resulted from the joining of words split over a line.  For example, the second most common case was 
\mbox{\<'r.Sy/sr'l> {\it erets-yisroel}} `land of Israel' with the final form \<.S> in the middle of the word.  Spot-checking a few revealed that they arose from 
\mbox{\<'r.S\textrm{-}y/sr'l>} split over a line, with a hyphen originally ending the line and being removed at some point. 

Other cases were genuine \ocr errors. The example above
in Section \ref{app:aspects:pointed}
with the pointed Yiddish characters is one such case, with the incorrect word-medial \<N> in the incorrect \ocr word.
There are a variety of other cases.  As another example, in one instance \mbox{\<'wN 'a>} {\it un a} `and a' was \ocr'd as 
\mbox{\<'y N'a>}.

\subsubsection{Low-frequency words}
\label{app:aspects:unusual:2}
We also carried out some cursory spot-checking of low-frequency words that seemed suspect.  For example, 
\<grwy``> {\it groyee} occurs 73 times, and checking the first such instance, the word in the source text is actually 
\<grwys`> {\it groyse} `big'.

\section{Tagset}
\label{app:tagset}

\begin{table*}
\center
{\small
    \begin{tabular}{|l|r||l|r||l|r|} \hline
tag & count & tag & count & tag & count \\ \hline
 PUNC       & 15,107  &     QR         &    46  &           RP-H\_S0    &     4  \\ \hline
 N          & 10,305  &     RP-ADJ     &    42  &           RP-H\_S1    &     4  \\ \hline
 PRO        &  7,981  &     P\textasciitilde{}WPRO     &    41  &           TO\textasciitilde{}VB      &     4  \\ \hline
 D          &  6,651  &     Q\_S0       &    39  &           WPRO\$      &     4  \\ \hline
 VBF        &  4,940  &     Q\_S1       &    39  &           C\textasciitilde{}NEG      &     3  \\ \hline
 ADV        &  4,193  &     ADVR       &    36  &           LS         &     3  \\ \hline
 P          &  3,813  &     RDN        &    35  &           NUM\_S3     &     3  \\ \hline
 NPR        &  2,706  &     NEG\_S0     &    34  &           N\_S0       &     3  \\ \hline
 VBN        &  2,603  &     NEG\_S1     &    34  &           N\_S1       &     3  \\ \hline
 CONJ       &  2,505  &     RP\textasciitilde{}TO\textasciitilde{}VB   &    33  &           P\textasciitilde{}N        &     3  \\ \hline
 VB         &  2,364  &     NPR\$       &    31  &          WADV\textasciitilde{}FP    &     3  \\ \hline
 HVF        &  2,084  &     RP-N       &    31  &           NPR\_S0     &     2  \\ \hline
 ADJ        &  2,055  &     X          &    29  &           NPR\_S1     &     2  \\ \hline
 MDF        &  1,663  &     VAG        &    28  &           NPR\_S2     &     2  \\ \hline
 C          &  1,414  &     VX         &    27  &           P\textasciitilde{}D\textasciitilde{}N      &     2  \\ \hline
 NEG        &  1,151  &     WADV\textasciitilde{}Q     &    27  &           ADJ\_S0     &     1  \\ \hline
 BEF        &  1,013  &     RP\textasciitilde{}VAN     &    25  &           ADJ\_S1     &     1  \\ \hline
 Q          &   936  &     WD         &    25  &           ADV-1      &     1  \\ \hline
 WPRO       &   864  &     FW         &    24  &           ADV\textasciitilde{}FP     &     1  \\ \hline
 INTJ       &   685  &     NUM\_S0     &    21  &           ADV\textasciitilde{}VAG    &     1  \\ \hline
 P\textasciitilde{}D        &   648  &     NUM\_S1     &    21  &           C\_S0       &     1  \\ \hline
 PRO\$       &   496  &    HVF\textasciitilde{}PRO    &    19  &           C\_S1       &     1  \\ \hline
 H          &   486  &     VBI\textasciitilde{}PRO    &    19  &           DR+P\_S0    &     1  \\ \hline
 VBI        &   456  &     ES\textasciitilde{}VBF     &    18  &           DR+P\_S1    &     1  \\ \hline
 RP\textasciitilde{}VBN     &   435  &     ADV\textasciitilde{}TO\textasciitilde{}VB  &    16  &           ES\textasciitilde{}BEF     &     1  \\ \hline
 NUM        &   395  &     BE         &    16  &           ES\textasciitilde{}HVF     &     1  \\ \hline
 RP         &   370  &     RD         &    15  &           FP\textasciitilde{}ADV     &     1  \\ \hline
 WADV       &   363  &     RP-V       &    15  &           FP\textasciitilde{}D       &     1  \\ \hline
 ES         &   252  &     WADV\_S0    &    15  &           H\_S2       &     1  \\ \hline
 RP\textasciitilde{}VB      &   249  &     WADV\_S1    &    15  &           MD-IPP     &     1  \\ \hline
 FP         &   245  &     BEF\textasciitilde{}PRO    &    13  &           MDF\textasciitilde{}ADJ    &     1  \\ \hline
 TO         &   245  &     BEN        &    13  &           MDF\textasciitilde{}NEG    &     1  \\ \hline
 MDN        &   221  &     WPRO\textasciitilde{}FP    &    13  &           MDF\textasciitilde{}VB     &     1  \\ \hline
 ADV\textasciitilde{}VBN    &   219  &     VB-DBL     &    12  &           NEG\textasciitilde{}VAG    &     1  \\ \hline
 RP-H       &   147  &     H\_S0       &    10  &           P-DBL\textasciitilde{}DR+P &     1  \\ \hline
 ADV\textasciitilde{}VB     &   140  &     H\_S1       &    10  &           P\textasciitilde{}NPR      &     1  \\ \hline
 VLF        &   133  &     NUM\_S2     &     9  &           P\textasciitilde{}PRO\textasciitilde{}VB   &     1  \\ \hline
 RP-ADV     &   107  &     ES\textasciitilde{}MDF     &     7  &           QTP        &     1  \\ \hline
 VBF\textasciitilde{}PRO    &   105  &     P-DBL      &     7  &           Q\textasciitilde{}D        &     1  \\ \hline
 DR+P       &    96  &     P\_S0       &     7  &           RP-N\textasciitilde{}VBN   &     1  \\ \hline
 ADV\_S0     &    93  &     P\_S1       &     7  &           RP\textasciitilde{}VAG     &     1  \\ \hline
 ADV\_S1     &    93  &     PRO\textasciitilde{}BEF    &     6  &           VB-1       &     1  \\ \hline
 MDF\textasciitilde{}PRO    &    76  &     RP-ADV\textasciitilde{}VB  &     6  &           VB-DBL-RSP &     1  \\ \hline
 PRO\textasciitilde{}VBF    &    75  &     ADV\textasciitilde{}VAN    &     5  &           VB-LFD     &     1  \\ \hline
 VAN        &    68  &     HVN        &     5  &           VBI\textasciitilde{}FP     &     1  \\ \hline
 PRO\textasciitilde{}HVF    &    59  &     PRO\textasciitilde{}VLF    &     5  &           WADVP-1    &     1  \\ \hline
 ADJR       &    56  &     RDF        &     5  &           WADVP-2    &     1  \\ \hline
 MD         &    56  &     D\textasciitilde{}N        &     4  &           WADV\textasciitilde{}MDF   &     1  \\ \hline
 N\$         &    52  &    INTJ\_S0    &     4  &           WPRO\_S0    &     1  \\ \hline
 PRO\textasciitilde{}MDF    &    51  &     INTJ\_S1    &     4  &           WPRO\_S1    &     1  \\ \hline
 ADJS       &    49  &     P\textasciitilde{}PRO      &     4  &           WPRO\textasciitilde{}PRO   &     1  \\ \hline
 NEG\textasciitilde{}ADV    &    49  &     RP-ADV\textasciitilde{}VBN &     4  &           {\bf total}      & 82,675  \\ \hline
\end{tabular}}
\caption{The complete tagset of 155 tags}
\label{app:tab:tagset}
\end{table*}

Table \ref{app:tab:tagset} shows the final tagset after the combining and splitting of words described
in Section \ref{sec:conversion:transform}.

\section{Conversion to Yiddish Script}
\label{app:conversion}
The default conversion for each word uses the \texttt{yiddish} package call:
\begin{verbatim}
yiddish.detransliterate(
  <word>, loshn_koydesh=True)
\end{verbatim}
where \texttt{<word>} is the romanized word.  The \texttt{loshn\_koydesh} flag indicates that it should convert to 
a Hebrew/Aramaic word if there is one corresponding to \texttt{<word>}.

In addition, for words with a hyphen, our wrapper converts  the individual components separately - e.g. (P\textasciitilde{}NPR \textit{far-peysekh}) `before Passover', with the results joined together.  In general (but not always, as mentioned below), the hyphenated words need to be split apart and converted separately to obtain the correct conversion.

\subsection{Handing special cases}
\label{app:conversion:special}
There are a variety of special cases, mostly resulting from minor discrepancies between the romanization in \ytb and what was required by \texttt{yiddish.detransliterate}.  See the code \texttt{convert\_to\_script.py} in the \texttt{ppchyprep} package mentioned in footnote \ref{fn:ppchyprep}
for complete details, but roughly:

\begin{enumerate}
    \item For some (word, pos) cases, the \texttt{loshn\_koydesh}
 flag is set to \texttt{False}, when the Hebrew word was not the one desired - e.g. (VBI \textit{shem}) (the example from Section \ref{sec:orthography}). 
 
 \item For some words with a hyphen, the hyphen needs to first be removed to return the correct conversion - e.g., (N \textit{eyn-hore}) `evil eye'.
 
 \item In other cases, words with a hyphen need to be passed in as a unit instead of the default of being done separately  - e.g. (N \textit{beys-hamigdesh})`Temple in Jerusalem'.
 
 \item Miscellaneous cases in which we hard-coded the correct conversion, without calling  \texttt{detransliterate}.
  A number of these cases include Hebrew words quoted in the material and not included in the \texttt{yiddish} conversion.
 
 \item There are a number of cases in which the romanization in the \ytb was not quite the same as what was expected by \texttt{yiddish}, and in such cases we simply modify the word before passing it to \texttt{detransliterate} - e.g. changing (RP-H \textit{maskim)}) to  (RP-H \textit{maskem)})
 \end{enumerate}

\subsection{Testing the conversion}
In earlier work, we had implemented our own conversion of the romanization to Yiddish script and  
tested the algorithm by comparing the converted version of \citet{grinefelder} to the original Yiddish script source text of the play.  While the exact edition used for the treebank is not part of the \ybc corpus, we used an earlier edition \citep{grinefelder2}.  We carried out a Smith-Waterman alignment between the words in the two versions and manually inspected it to verify the correctness of the conversion.  This procedure caught a number of the non-phonetic words that had not been converted properly.  

While we are now using the \texttt{yiddish} package instead of that earlier conversion code, we compared all cases where the \texttt{yiddish} package gave different results from our earlier conversion code. While in some cases the new conversion was correct, that was not always the case, and this process identified a number of the cases that were then handled as discussed in Section \ref{app:conversion:special}.   In future work we will again directly compare the converted version to the Yiddish script source text using Smith-Waterman.

 The other text that we are using, from \citet{olsvanger}, lacks an original source text in Yiddish script, since the source text was written in a romanized dialect representation.  We are in effect converting it to a non-existent Yiddish script source.  In the earlier version, we tested the conversion by reasoning that if a non-phonetic word was incorrectly converted from the romanized form to Yiddish script, the resulting ``word'' would likely never appear in the \ybc corpus, and so we checked whether each  converted word exists in the \ybc corpus.  In future work we can test the current conversion in the same way.\footnote{A version of this material in Yiddish script has recently appeared \citep{olsvanger2022} - \url{https://forward.com/yiddish/548219}.  The same procedure as with \citet{grinefelder} could therefore be carried out, but only with an electronic copy of the material.}

\subsection{Alternate converted forms}

The \texttt{yiddish} library converts the words to a \svo form. As discussed in Section \ref{sec:orthography}, the words in the \ybc corpus have a great deal of spelling variation and do not necessarily follow the \svo.  In addition to the sort of spelling variations mentioned in that section, they differ on the use of diacritics.  

We therefore created two alternate forms of the words from the conversion of the \ytb files for use in training and evaluation of the \pos tagger.  The first alternate form results from replacing  \<yaya> (pasekh tsvey yudn ) with  \<yy> (tsvey yudn).  The second alternate form results from also replacing \<'a> (paskeh alef) with 
\<'> (shtumer alef) and \<yi> (khirek yud) with \<y> (yud).

We do not discuss the results with these alternate forms in this paper, but they did not improve the current results.  However, as discussed in Section \ref{sec:results:limitations} and the conclusion, our evaluation data is currently very limited, and we will revisit the use of these alternate forms when evaluating the tagger with material from the \ybc and elsewhere.  The software in \texttt{ppchyprep} contains the code for producing these alternate forms.

\section{Contextualized Embeddings Trained on the YBC}
\label{app:con_embeddings}

\begin{table}[t]
    \centering
    {\small
    \begin{tabular}{|l|r|} \hline
     parameter & value \\ \hline
     model\_name\_or\_path &  xlm-roberta-base \\ \hline
     fp16 & True \\ \hline
     per\_device\_train\_batch\_size & 16 \\ \hline
     gradient\_accumulation\_steps & 16 \\ \hline
     per\_device\_eval\_batch\_size & 16 \\ \hline
    \end{tabular}}
    \caption{Parameters used for continued pre-training of \xlmrobertabase. These were the input parameters to the run\_mlm.py script in the hugging face transformers package.}
    \label{tab:pretraining}
\end{table}

The training consisted of 10 epochs of continued pretraining on \xlmrobertabase.   Since it was continued pretraining, it used the existing configuration for \xlmrobertabase, including the tokenization.  The parameters used for the continued pretraining are in Table \ref{tab:pretraining}.

Of the 9805 files in the \ybc corpus, 1\% were randomly put into the validation section.  The resulting train and validation section sizes are shown in Table \ref{app:ybcsplit}

\begin{table}
\centering
{\small
    \begin{tabular}{|c|r|r|}  \hline
    section & \# files & \#tokens \\ \hline
    train & 9714 & 647,885,718 \\ \hline
    val & 91 & 5,440,472 \\ \hline
    total & 9805 & 653,326,190 \\ \hline
    \end{tabular}}
    \caption{Split of the \ybc corpus for training embeddings}
\label{app:ybcsplit}
\end{table}

\section{Non-contextualized Embeddings Trained on the YBC}
\label{app:noncon_embeddings}

\begin{table}[t]
    \centering
    {\small
    \begin{tabular}{|l|r|} \hline
         parameter & value \\ \hline
        memory& 100.0 \\ \hline
        min\_count  & 5 \\ \hline
        vector\_size & 300 \\ \hline
        iter M & 25 \\ \hline
        window\_size  & 10 \\ \hline
        binary & 2 \\ \hline
        x\_max  & 10 \\ \hline
    \end{tabular}}
    \caption{Parameters used for training \glove embeddings}
    \label{tab:glove}
\end{table}

The \glove embeddings were trained using version 1.2 of \glove, with the parameters shown in Table \ref{tab:glove}.  The same train/val split of the \ybc corpus was used as for the contextualized embeddings. 

\section{POS Tagger Training}
\label{app:model}

\begin{table*}[t]
    \begin{subtable}{.45\linewidth}
    \centering
    {\small
    \begin{tabular}{|l|l|} \hline
         parameter & value \\ \hline
         \multicolumn{2}{|c|}{\it TransformerWordEmbeddings} \\ \hline
         layers & -1 \\ \hline
         subtoken\_pooling & first \\  \hline
         fine\_tune & True \\ \hline
         \multicolumn{2}{|c|}{\it SequenceTagger} \\ \hline    
        hidden\_size & 256 \\ \hline
        tag\_type & pos \\ \hline
        use\_crf & True \\ \hline
        use\_rnn & False \\ \hline
        reproject\_embeddings & False \\ \hline
    \multicolumn{2}{|c|}{\it ModelTrainer} \\ \hline
    learning\_rate & 5e-5 \\ \hline
    mini\_batch\_size  & 32 \\ \hline
    mini\_batch\_chunk\_size & 1 \\ \hline
    max\_epochs & 50 \\ \hline
    weight\_decay & 0.0 \\ \hline
    optimizer  & torch.optim.AdamW \\ \hline
    scheduler & LinearSchedulerWithWarmup \\ \hline
    warmup\_fraction & 0.1 \\ \hline
    use\_final\_model\_for\_eval & False \\ \hline
    \end{tabular}}
    \caption{Using \xlmrobertabase embeddings}
    \label{tab:flair1}
\end{subtable}
\hfill
\begin{subtable}{.45\linewidth}
    \centering
    {\small
    \begin{tabular}{|l|l|} \hline
         parameter & value \\ \hline
         \multicolumn{2}{|c|}{\it WordEmbeddings} \\ \hline
         \multicolumn{2}{|c|}{(all default)} \\ \hline
         \multicolumn{2}{|c|}{\it SequenceTagger} \\ \hline    
        hidden\_size & 256 \\ \hline
        tag\_type & pos \\ \hline
        use\_crf & True \\ \hline
        use\_rnn & default (True) \\ \hline
        reproject\_embeddings & default (True) \\ \hline
    \multicolumn{2}{|c|}{\it ModelTrainer} \\ \hline
    learning\_rate & 0.1 \\ \hline
    mini\_batch\_size  & 32 \\ \hline
    mini\_batch\_chunk\_size & 1 \\ \hline
    max\_epochs & 50 \\ \hline 
    use\_final\_model\_for\_eval & False \\ \hline
    \end{tabular}}
    \caption{Using \glove embeddings}
    \label{tab:flair2}
\end{subtable}
\caption{Parameters used for training \pos tagger}
\end{table*}

The \pos models were trained with version 0.12.2 of \flair, with a slightly modified form of their supplied \texttt{run\_ner.py} script.  The parameters used for training the \pos models with \xlmrobertabase embeddings are shown in Table \ref{tab:flair1}, and those  used for training the models using \glove embeddings are shown in Table \ref{tab:flair2}. 

With \texttt{use\_final\_model\_for\_eval} set to \texttt{False}, the validation section was used to save the model with the best performance on that section.  In both cases we omit parameters referring to the name or location of the embeddings, output directory, etc.

\end{document}